%% file: naaclhlt2019.tex
\documentclass[11pt,a4paper]{article}
\usepackage[hyperref]{naaclhlt2019}
\usepackage{times}
\usepackage{latexsym}

\input{macro.tex}

\aclfinalcopy 

\title{What Makes Good In-Context Examples for GPT-\boldsymbol{$3$}?}

\author{\makecell{Jiachang Liu$^{1}$\thanks{\hspace{.06in}Work was done during an internship at Microsoft Dynamics 365 AI.}, Dinghan Shen$^{2}$, Yizhe Zhang$^{3}$, Bill Dolan$^{3}$, Lawrence Carin$^{1}$, Weizhu Chen$^2$} \\
$^1$Duke University \quad
$^2$Microsoft Dynamics 365 AI \quad
$^3$Microsoft Research\\
$^1$\{jiachang.liu, lcarin\}@duke.edu \\ $^{2,3}$\{dishen, yizzhang, billdol, wzchen\}@microsoft.com} 

\date{}

\begin{document}
\maketitle
\begin{abstract}
GPT-$3$ \cite{brown2020language} has attracted lots of attention due to its superior performance across a wide range of NLP tasks, especially with its powerful and versatile in-context few-shot learning ability.
Despite its success, we found that the empirical results of GPT-$3$ depend heavily on the choice of in-context examples.
In this work, we investigate whether there are more effective strategies for judiciously selecting in-context examples (relative to random sampling) that better leverage GPT-$3$'s few-shot capabilities.
Inspired by the recent success of leveraging a retrieval module to augment large-scale neural network models, we propose to retrieve examples that are semantically-similar to a test sample to formulate its corresponding prompt.  
Intuitively, the in-context examples selected with such a strategy may serve as more informative inputs to unleash GPT-$3$'s extensive knowledge.
We evaluate the proposed approach on several natural language understanding and generation benchmarks, where the retrieval-based prompt selection approach  consistently outperforms the random baseline. 
Moreover, it is observed that the sentence encoders fine-tuned on task-related datasets yield even more helpful retrieval results.
Notably, significant gains are observed on tasks such as table-to-text generation (41.9\% on the ToTTo dataset) and open-domain question answering (45.5\% on the NQ dataset).
We hope our investigation could help understand the behaviors of GPT-$3$ and large-scale pre-trained LMs in general and enhance their few-shot capabilities.
\end{abstract}

\section{Introduction}
GPT-$3$ \cite{brown2020language} is a new breakthrough in NLP research.
Previously, NLP models are pre-trained on large quantities of data and fine-tuned on a specific task and dataset.
What sets GPT-$3$ apart from other pre-trained language models is its impressive ``in-context'' few-shot learning ability.
Provided with a few in-context examples, GPT-$3$ is able to generalize to unseen cases without further fine-tuning.
This opens up many new technological possibilities that are previously considered unique to human. 
For example, NLP systems can be developed to expand emails, extract entities from text, generate code based on natural language instructions with a few demonstration examples.

Despite its powerful and versatile in-context learning ability, GPT-$3$ has some practical challenges/ambiguities. 
The original paper~\cite{brown2020language} utilizes task-relevant examples that are randomly sampled from the training set to construct the context.
In practice, we observe that the performance of GPT-$3$ tends to fluctuate with different choices of in-context examples.
As shown in Table~\ref{tab:GPT3_unstable_results}, the variance of the empirical results with distinct in-context examples can be significant.
The results are highly sensitive to the examples.
Our work aims to carefully examine this issue to gain a deeper understanding on how to better select in-context examples to unleash GPT-$3$'s few-shot capabilities and further improve its performance.

\begin{table}
\centering
\begin{small}
\setlength{\tabcolsep}{5pt}
\def\arraystretch{1.05}
\begin{tabular}{@{}c||ccccc@{}}
\toprule[1.2pt]
\textbf{Trial}    & 1    & 2    & 3    & 4    & 5    \\ \hline
\textbf{Accuracy} & 94.6 & 95.0 & 95.8 & 93.9 & 86.9 \\
\bottomrule[1.2pt]
\end{tabular}%
\caption{Results of GPT-$3$ on the task of sentiment analysis on the SST-2 dataset. Five different in-context examples are randomly selected from the training set. We observe different contexts induce different accuracies on the test set.}
\label{tab:GPT3_unstable_results}
\end{small}
\vspace{-3mm}
\end{table}

A brute-force approach would be to perform combinatorial search over the entire dataset. Unfortunately, this strategy is computationally expensive and thus impractical in many cases.
To this end, we investigate the influences of employing different in-context examples on the empirical results. Interestingly, we found that the in-context examples that are closer to the test sample in the embedding space consistently give rise to stronger performance (relative to the farther ones).
Inspired by this observation and the recent success of retrieval-augmented models~\cite{hashimoto2018retrieve},
we propose to utilize nearest neighbors of a given test sample (among all the training instances available) as the corresponding in-context examples. The retrieved examples, along with the test sample, are provided to GPT-$3$ for the final prediction.

To verify the effectiveness of the proposed method, we evaluate it on several natural language understanding and generation tasks, including sentiment analysis, table-to-text generation and open-domain question answering.
It is observed that the retrieval-based in-context examples unleash the few-shot capabilities of GPT-$3$ much more effectively than a random sampling baseline. 
Even with a smaller number of in-context examples, the proposed strategy empowers GPT-$3$ to achieve stronger performance.
Moreover, we find that the specific sentence encoders employed for the retrieval procedure play a critical role. Thus, an extensive exploration regarding different pre-trained encoders is conducted, and it is shown that encoders fine-tuned on natural language matching tasks serve as more effective in-context examples selector on the QA task.
Detailed analysis and case study further validate the effectiveness of proposed methods. In summary, our contributions in this
paper are as follows:

\emph{\romannumeral1}) to  the  best  of  our  knowledge, we take a first step towards understanding the sensitivity of GPT-$3$'s few-shot capabilities with respect to the selection of in-context examples;

\emph{\romannumeral2}) to alleviate the sensitivity issue, an additional retrieval module is introduced to find semantically-similar in-context examples of a test instance to construct its corresponding input, which greatly outperforms the baseline based on random sampled examples;

\emph{\romannumeral3}) fine-tuning the retrieval model on task-related dataset(s) leads to even stronger empirical results with GPT-$3$;

\emph{\romannumeral4}) the performance of GPT-$3$ improves as the number of examples available for retrieval increases.

\section{Method}
\vspace{-1mm}
\subsection{GPT-$3$ for In-Context Learning}
\begin{figure}
    \centering
    \includegraphics[width=\columnwidth]{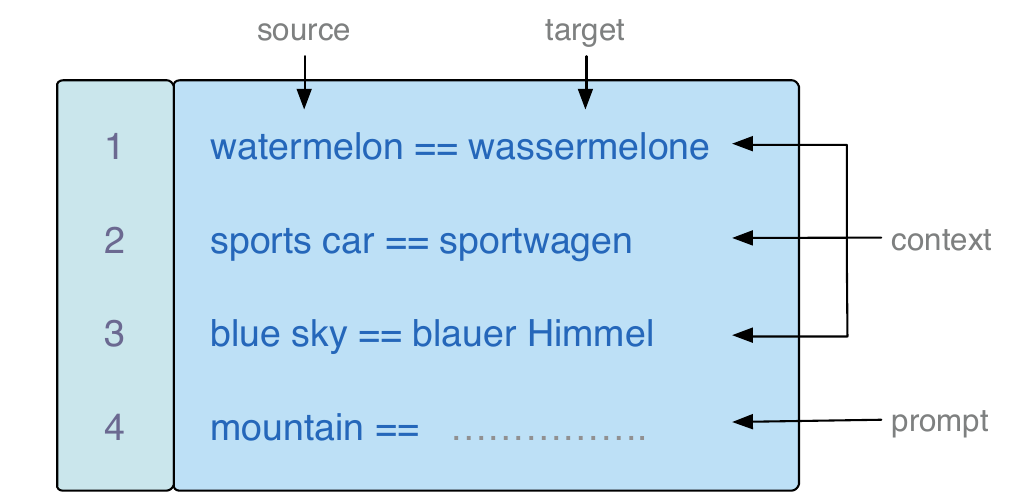}
    \caption{The figure above shows how to perform in-context learning with a language model. Three in-context examples and the test prompt are concatenated as a single string input for GPT-$3$, with a special character "\textbackslash{}n" inserted between two adjacent examples. GPT-$3$ keeps generating tokens until there is a special character "\textbackslash{}n".}
    \label{fig:in-context_learning_diagram}
\vspace{-3mm}
\end{figure}

\vspace{-1mm}
The in-context learning scenario of GPT-$3$ can be regarded as a conditional text generation problem.
Concretely, the probability of generating a target $y$ is conditioned on the context $C$, which includes $k$ examples, and the source $x$. Therefore, the prediction $y$ corresponding to the source $x$ can be expressed as:
\begin{equation}
    p_{\text{LM}}(y|C, x) = \prod_{t=1}^T p(y_t|C, x, y_{<t})
\end{equation}
where $\text{LM}$ denotes the parameters of the language model, and $C=\{x_1, y_1, x_2, y_2, ..., x_k, y_k\}$ is a context string. In GPT-$3$, the $C$ is created by concatenating $k$ training instances along with their corresponding labels.
As shown in the illustration of Figure~\ref{fig:in-context_learning_diagram}, GPT-$3$ is asked to translate ``mountain'' to its German version based on the three examples given as part of the input.

For GPT-$3$, this generation process is implemented through a giant transformer-based model architecture~\cite{vaswani2017attention, brown2020language}.
Given the large size of the GPT-3 model, it would be very computationally-involved to fine-tune it on task-specific samples.
Thus, GPT-$3$ is typically leveraged in a in-context learning manner as described above.
It has been shown that GPT-$3$ has powerful few-shot capabilities, where it can perform quite well with only a small number of demonstrations provided. 
Unfortunately, as shown in Table~\ref{tab:GPT3_unstable_results}, the results of GPT tends to fluctuate significantly with different in-context examples chosen.
Here we aim to alleviate this issue via judious in-context examples selection.

\begin{figure*}[t]
    \centering
    \includegraphics[width=\textwidth]{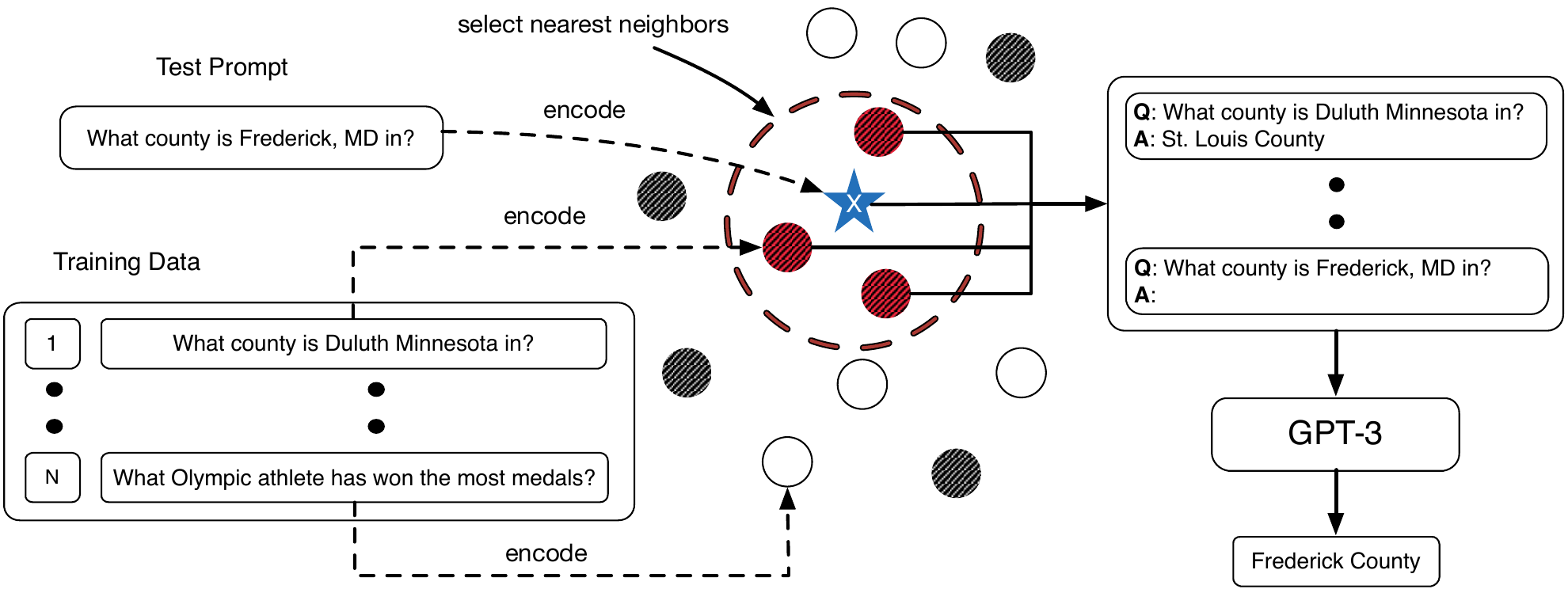}
    \caption{In-context example selection for GPT-$3$. White dots: unused training samples; grey dots: randomly sampled training samples; red dots: training samples selected by the $k$-nearest neighbors algorithm in the embedding space of a sentence encoder.}
    \label{fig:knn_workflow}
\end{figure*}

\subsection{The Impact of In-Context Examples}
\label{sec:connection_to_VAE}
Given the observation that the empirical results of GPT-$3$ are sensitive to the chosen in-context examples, we look at the role of in-context examples from an empirical perspective.
Previous retrieve-and-edit literatures usually retrieve prototypes that are close to the test source $x$ in some embedding space.
These examples and the test source $x$ often share semantic or lexical similarities.
This hints on how we may select in-context examples for GPT-$3$.

To this end, we examine the impact of the distance between in-context example and the test sample on GPT-$3$'s performance. Concretely, a comparison is made on the the Natural Questions (NQ) dataset between two in-context example selection strategies. Given each test example, the first method utilizes the $10$ farthest training instances to construct the context provided to GPT-$3$, while the second employs the  $10$ closest neighbors. 
We use the CLS embeddings of a pre-trained RoBERTa-large model as the sentence representations to measure the proximity of two sentences (using the Euclidean distance).
\begin{table}[b]
\centering
\begin{small}
\setlength{\tabcolsep}{7pt}
\def\arraystretch{1.05}
\begin{tabular}{@{}c||cc@{}}
\toprule[1.2pt]
\textbf{Method}   & Closest         & Farthest  \\ \hline
\textbf{Accuracy} & \textbf{46.0} & 31.0 \\
\bottomrule[1.2pt]
\end{tabular}%
\caption{Comparison of the EM score on the closest 10 neighbors and farthest 10 neighbors on a subset of 100 test samples of the NQ dataset.}
\label{tab:close_far_comparison}
\end{small}
\vspace{-3mm}
\end{table}

For evaluation, $100$ test questions are randomly sampled and the average Exact Match (EM) scores with the two distinct strategies are reported in Table~\ref{tab:close_far_comparison}.
It can be observed that the nearest neighbors, as the in-context examples, give rise to much better results relative to the farthest ones. Moreover, the pre-trained RoBERTa model serves as effective sentence embeddings for the retrieval procedure.

\subsection{$k$NN-augmented In-Context Example Selection}
\label{sec:KATE_proposal_method}
Based on the findings above, we propose KATE\footnote{KATE: \textbf{K}nn-\textbf{A}ugmented in-con\textbf{T}ext \textbf{E}xample selection}, a strategy to select good in-context examples for in-context learning. The process is visualized in Figure~\ref{fig:knn_workflow}. Specifically, we first use a certain sentence encoder to convert sources in both the training set and test set to vector representations. For online prediction, we can convert the training set first and encode each test source on the fly. Then, for each test source $x$, we retrieve its nearest $k$ neighbors $x_1, x_2, ..., x_k$ from the training set (according to the distances in the sentence encoder's embedding space). Given some pre-defined similarity measure $d$ such as the cosine similarity,  the neighbors are ordered in such a way that $d(x_i, x) \leq d(x_j, x)$ when $i < j$.

\begin{algorithm}
\caption{$k$NN In-context Example Selection}
\textbf{Given:} test prompt $\vx_{\text{test}}$, training set $\mathcal{D}_T=\{\vx_i, \vy_i\}_{i=1}^N$, sentence encoder $\mu_{\theta}(\cdot)$, and number of in-context examples $k$ (hyperparameter).
\begin{algorithmic}[1]
    \State $\vv_{\text{test}} = \mu_{\theta}(\vx_{\text{test}})$
    \For{$\vx_i \in \mathcal{D}_T$}
        \State $\vv_i = \mu_{\theta}(\vx_i)$
        \State $\vs_i = -\lVert \vv_{\text{test}} - \vv_i \rVert_2$ \text{ } \text{ } \text{ } (or $\frac{\vv_{\text{test}} \cdot \vv_i}{\lVert \vv_{\text{test}} \rVert_2 \lVert \vv_i \rVert_2}$)
    \EndFor
    \State Select largest $k$ similarities $\vs_i$'s (in descending order) with indices $\{\sigma(1), ..., \sigma(k)\}$
    \State $C = [\vx_{\sigma(1)}; \vy_{\sigma(1)}; ...; \vx_{\sigma(k)}; \vy_{\sigma(k)}]$
    \State $\hat{\vy}_{\text{test}} = \text{GPT-}3 ([C; \vx_{\text{test}}])$
\end{algorithmic}
 \label{alg:knn}
\end{algorithm}

Afterwards, the $k$ sources are concatenated with their corresponding targets to form the context $C=\{x_1, y_1, x_2, y_2, ..., x_k, y_k\}$, which is further sent to GPT-$3$ along with the test input. The algorithm chart is presented in Algorithm~\ref{alg:knn}. Note that different numbers of in-context examples can be employed here, and we conduct ablation study on its impact in a later section.

\paragraph{Choices of Retrieval Module}
A core step for our context selection approach is mapping sentences into a latent semantic space, leaving a question as what sentence encoders we should choose.
We compared among existing pre-trained text encoders and found them  sufficient to retrieve semantically similar sentences.
The sentence encoders can be divided into two categories.

The first category includes most generally pre-trained sentence encoders such as a pre-trained BERT, RoBERTa, or XLNet models.
These models have been trained on large quantities of unsupervised tasks and achieved good performance on many natural language tasks.
The corresponding embeddings contain rich semantic information from the original sentences.

The second category includes sentence encoders fine-tuned on specific tasks or datasets.
For example, a sentence encoder trained on the STS benchmark dataset should be able to assess similarities among different questions better than a generally pre-trained sentence encoder.
\cite{reimers2019sentence, reimers2020making} have shown that these fine-tuned encoders have achieved great performance on tasks such as sentence clustering, paraphrase mining, and information retrieval.

\section{Experimental Setup}

\begin{table}
\centering
\begin{small}
\setlength{\tabcolsep}{8pt}
\def\arraystretch{1.05}
\begin{tabular}{@{}c||ccc@{}}
\toprule[1.2pt]
\textbf{Dataset} & \textbf{Train} & \textbf{Dev} & \textbf{Test} \\ \hline
SST-2            & 67k            & 872          & 1.8k          \\
IMDB             & 25k            & -            & 25k           \\
ToTTo            & 120k           & 7.7k         & 7.7k          \\
NQ               & 79k            & 8.8k         & 3.6k          \\
WQ               & 3.4k           & 361          & 2k            \\
TriviaQA         & 78.8k          & 8.8k         & 11.3k         \\ \bottomrule[1.2pt]
\end{tabular}%
\caption{Data split for different datasets. In-context examples are selected from the training set. Because ToTTo and TriviaQA require submitting to their leaderboards, the evaluation is done on the dev sets. For all other datasets, the evaluation is done on the test sets.}
\label{tab:dataset_split}
\end{small}
\end{table}

We apply the $k$NN in-context selection method to the following three tasks: sentiment classification, table-to-text generation, and question answering (QA).
Datasets and the common data split setups are shown in Table~\ref{tab:dataset_split}.  
In terms of the hyper-parameters in the GPT-$3$ API, we set the temperature to 0. We let GPT-$3$ keep generating tokens until there is a special token ``\textbackslash{}n".

\label{sec:list_sentence_encoder}

\subsection{Sentence Embeddings for Retrieval}
To retrieve semantically-similar training instances, we consider two types of sentence embeddings.
\begin{itemize}
    \item The original pre-trained RoBERTa-large model~\cite{liu2019roberta}, which is abbreviated as KATE$_{\text{roberta}}$;
    \item The RoBERTa-large model fine-tuned on task-related datasets: \emph{\romannumeral1}) fine-tuned on the SNLI and MultiNLI dataset (KATE$_{\text{nli}}$); \emph{\romannumeral2}) first fine-tuned on the SNLI and MultiNLI dataset and then on the STS-B dataset (KATE$_{\text{nli+sts-b}}$).
\end{itemize}
Notably, all the sentence encoders share the same the architecture, where the only differences are the specific datasets used for fine-tuning.
Euclidean distance is used for the KATE$_{\text{roberta}}$ case, while cosine similarity is employed for KATE$_{\text{nli}}$ and KATE$_{\text{nli+sts-b}}$.

\paragraph{Sentiment Analysis}
For sentiment classification, we select in-context examples under the transfer setting, where one dataset is treated as the training set and the evaluation is made on another dataset. 
This transfer setting is designed to simulate a real-world scenario where 
we would like to leverage an existing labeled dataset for a unlabeled one (of a similar task). 

Specifically, we select in-context examples from the SST-2 training set~\cite{socher2013recursive, wang2018glue} and ask GPT-$3$ to make predictions on the IMDB test set~\cite{maas2011learning}.
To explore whether a sentence encoder fine-tuned on a similar task would benefit KATE's performance, we also employ a pre-trained RoBERTa-large model fine-tuned on the SST-2 training set (dubbed as KATE$_{\text{sst-2}}$).
The performance is measured by the accuracy over the entire IMDB test set.
The number of in-context examples is chosen to be $3$ since adding more examples does not further improve the performance. 

\paragraph{Table-to-Text Generation}
Given a Wikipedia table and a set of highlighted cells, this task focuses on producing human-readable texts as descriptions. 
ToTTo~\cite{parikh2020totto} is utilized for evaluation due to its popularity.
We use BLEU~\cite{papineni2002bleu} and PARENT~\cite{dhingra2019handling}. metrics for evaluation.
The ToTTo code base contains both evaluation and preprocessing scripts\footnote{The ToTTo code base can be found at \url{https://github.com/google-research/language/tree/master/language/totto}}.
Due to the input length limit of GPT-$3$ (currently the token limit is 2048), we add an extra preprocessing step by deleting the closing angle brackets such as $<$/cell$>$ and $<$/table$>$ to save some space.
The number of in-context examples is set as $2$ .

\paragraph{Question Answering}
Given a factual question, the model is asked to generate the correct answer.
Following prior studies, we use the Exact Match (EM) score to measure the performance of GPT-$3$ on open-domain QA tasks.
The EM score is defined as the proportion of the number of predicted answers being exactly the same as (one of) the ground-truth answer(s).
The matching is performed after string normalization, which includes article and punctuation removal.
We conduct experiments on three open-domain QA benchmarks: Natural Questions (NQ)~\cite{kwiatkowski2019natural}, Web Questions (WQ)~\cite{berant2013semantic}, and Trivia Question Answering (TriviaQA)~\cite{joshi2017triviaqa}.
For this task, we pick the nearest 64 neighbors as the in-context examples for NQ and WQ and nearest 10 neighbors for TriviaQA (The retrieved 64 examples could not fit into 2048 token limit for TriviaQA. For fair comparison, we set the number of in-context examples to be 10 for TriviaQA for both the baseline and KATE method).
The evaluation is done on the test sets of NQ and WQ and the dev set of TriviaQA.
\subsection{Baseline Methods}
\label{sec:baseline_method}
\paragraph{Random Sampling} For each test sentence, we randomly select in-context examples from the training set.
We refer to this method as \textit{Random} in the experimental results.
To have a fair comparison with KATE, the number of in-context examples in this random baseline is the same as KATE to ensure fair comparison.
On the test set, the random baseline is repeated for five times to obtain the average score and corresponding standard deviation.

\paragraph{$k$-Nearest Neighbor} Additionally, to investigate whether the retrieval module is complementary to GPT-$3$'s few-shot learning ability, we further consider a $k$-nearest neighbor baseline.
Specifically, for text generation tasks, the target $y_1$ associated with the first retrieved example is considered as the predicted target for the test sample. 
As to the sentiment analysis and QA tasks, the top $k$ retrieved examples $\{y_1, ..., y_k\}$ are utilized, where the final prediction is determined by majority voting among the $k$ examples' targets. If there is a tie case, we take the target of the example that is most similar to the test sentence as the prediction.
To ensure fair comparison, we compare the baseline $k$NN and KATE under the same embedding space of a pre-trained RoBERTa-large model.
This baseline is abbreviated as $k$NN$_{\text{roberta}}$.

\section{Experimental Results}

\subsection{Sentiment Analysis}
\begin{table}[t]
\centering
\begin{small}
\setlength{\tabcolsep}{10pt}
\def\arraystretch{1.05}
\begin{tabular}{@{}c||c@{}}
\toprule[1.2pt]
\textbf{Method} & \textbf{Accuracy} \\ \hline
Random                         & 87.95 $\pm$ 2.74 \\
$k$NN$_\text{roberta}$ & 50.20 \\
KATE$_{\text{roberta}}$                  & 91.99            \\
KATE$_{\text{nli}}$                  & 90.40            \\
KATE$_{\text{nli+sts-b}}$                  & 90.20            \\
KATE$_{\text{sst-2}}$ & \textbf{93.43} \\
\bottomrule[1.2pt]
\end{tabular}%
\caption{Accuracy of sentiment prediction for GPT-$3$ on IMDB with different choices of in-context examples. In-context examples are from the SST-2 dataset.}
\label{tab:SST-2_to_IMDB}
\end{small}
\end{table}
\vspace{0mm}

We first evaluate KATE on the sentiment analysis task.
The results are shown in Table~\ref{tab:SST-2_to_IMDB}.
It can be observed that KATE consistently produces better performance relative to the random selection baseline. 
Notably, there is no variance with the obtained results since the same set of retrieved in-context examples are employed.
For the KATE method, when a pre-trained sentence encoder is fine-tuned on NLI or NLI+STS-B datasets, the performance slightly decreases.
Since the objectives of the IMDB dataset and the NLI+STS-B datasets are different, this shows that fine-tuning on a dissimilar task can hurt KATE's performance.
Moreover, KATE$_{\text{nli+sts-b}}$ performs worse than KATE$_{\text{nli}}$ because the sentence encoder has been further fine-tuned on the STS-B dataset.
In contrast, KATE$_{\text{sst-2}}$ obtains the best accuracy, showing that fine-tuning on a similar task can benefit KATE's performance.
To verify that the gains are not merely from the retrieval step, we further compare KATE$_{\text{roberta}}$ with the $k$NN$_{\text{roberta}}$.
It turns out that the performance of the $k$NN$_{\text{roberta}}$ method is similar to random guessing.
This observation is consistent when one neighbor or three neighbors are retrieved.
Notably, with the embeddings of the RoBERTa-large model fine-tuned on the SST-2 dataset, the accuracy of $k$NN$_{\text{sst-2}}$ is 92.46, which is lower than that obtained with KATE$_{\text{sst-2}}$.
These results suggest that the GPT-3 model is critical to the final results, and the retrieval module is complementary to GPT-$3$'s few-shot capabilities.

\begin{table*}[th]
\centering
\begin{small}
\setlength{\tabcolsep}{5pt}
\def\arraystretch{1.05}
\begin{tabular}{@{}c||cccccc@{}}
\toprule[1.2pt]
\multirow{2}{*}{\textbf{Method}} & \multicolumn{2}{c}{\textbf{Overall}}     & \multicolumn{2}{c}{\textbf{Overlap Subset}} & \multicolumn{2}{c}{\textbf{Nonoverlap Subset}} \\ 
                       & \textbf{BLEU} & \textbf{PARENT} & \textbf{BLEU}   & \textbf{PARENT}  & \textbf{BLEU}    & \textbf{PARENT}    \\ \hline
Random                    & 28.4 $\pm$ 2.1 & 39.3 $\pm$ 2.6 & 31.2 $\pm$ 2.5 & 41.8 $\pm$ 3.0 & 25.6 $\pm$ 1.8 & 37.0 $\pm$ 2.3 \\ 
$k$NN$_{\text{roberta}}$                    & 14.1 & 12.6 & 20.1 & 17.9 & 8.0 & 7.52 \\
KATE$_{\text{roberta}}$                & \textbf{40.3} & \textbf{49.7} & \textbf{47.8} & \textbf{55.0} & \textbf{32.9} & \textbf{44.6} \\
KATE$_{\text{nli}}$                & 39.1 & 48.5 & 46.5 & 53.7 & 31.9 & 43.6 \\
KATE$_{\text{nli+sts-b}}$                & 38.1 & 47.2 & 45.2 & 52.2 & 31.1 & 42.4 \\
\bottomrule[1.2pt]
\end{tabular}%
\caption{Table-to-text generation results on the ToTTo dev dataset.}
\label{tab:totto_result}
\end{small}
\end{table*}
\input{ToTTo_retrieval_eg}

\vspace{-3mm}
\subsection{Table-to-text Generation}

We utilize the ToTTo dataset to evaluate KATE on the table-to-text generation task. The results are shown in Table~\ref{tab:totto_result}.
The KATE method gives rise to considerable gains over the random baseline, according to both the BLEU and PARENT scores.
On a finer scale, the evaluation can be done on the overlap subset and the nonoverlap subset. The overlap dev subset shares a significant number of header names with the training set, while the nonoverlap one does not share any header names.
It can be observed that the KATE method improves the results on both the overlap and the nonoverlap subsets, meaning that the retrieval module is helpful for both situations where the test set follows the distribution of the training set and where the test set is out of distribution of the training set.
Similar to sentiment analysis, there is a slight drop in performance from KATE$_{\text{roberta}}$ to KATE$_{\text{nli}}$ and KATE$_{\text{nli+sts-b}}$.
This is due to the difference between the objectives of the ToTTo dataset and NLI+STS-B datasets.
The drop from KATE$_{\text{nli}}$ to KATE$_{\text{nli+sts-b}}$ further validates the idea that fine-tuning on a dissimilar task can hurt KATE's performance.
For the $k$NN baseline, it performs much worse than the random selection method and the KATE method, again suggesting that the retrieval process and GPT-3 work together collaboratively to achieve better results.

To understand how the retrieval mechanism helps GPT-$3$'s predictions, we conduct a case study on the retrieved examples (see Table~\ref{tab:ToTTo_retrieval_eg}). By retrieving relevant examples from the training set, KATE provides useful detailed information within the table, \emph{e.g.}, the number of points, rebounds, and assists, to GPT-$3$ for more accurate description. On the other hand, the random selection method has the issue of hallucination, where the generated sequences contain information (\emph{i.e.}, ``senior year'' and ``University of Texas'') not present in the table. 

\begin{table}[t]
\centering
\begin{small}
\setlength{\tabcolsep}{3pt}
\def\arraystretch{1.05}
\begin{tabular}{@{}c||ccc@{}}
\toprule[1.2pt]
\textbf{Method} & \textbf{NQ} & \textbf{WQ} & \textbf{TriviaQA}$^*$ \\ \hline
RAG (Open-Domain)       & 44.5        & 45.5        & 68.0              \\
T5+SSM (Closed-Book)    & 36.6        & 44.7        & 60.5              \\
T5 (Closed-Book)        & 34.5        & 37.4        & 50.1              \\
GPT-$3$ (64 examples)                 & 29.9        & 41.5        & -                 \\ \hline
\multicolumn{4}{c}{Ours}                                              \\ \hline
Random                               & 28.6 $\pm$ 0.3   & 41.0 $\pm$ 0.5   & 59.2 $\pm$ 0.4         \\
$k$NN$_{\text{roberta}}$                              & 24.0   & 23.9  &  26.2        \\
KATE$_{\text{roberta}}$           & 40.0        & 47.7        & 57.5              \\
KATE$_{\text{nli}}$           & 40.8        & \textbf{50.6}        & 60.9              \\
KATE$_{\text{nli+sts-b}}$           & \textbf{41.6}        & 50.2        & \textbf{62.4}              \\
\bottomrule[1.2pt]
\end{tabular}%
\caption{QA results on various datasets. (*) On TriviaQA, we used 10 examples. On NQ and WQ, we used 64 examples.}
\label{tab:QA_result}
\end{small}
\end{table}
\vspace{-3mm}

\subsection{Questing Answering}

\input{NQ_retrieval_eg}

\begin{figure*}[ht]
    \centering
    \includegraphics[width=\columnwidth]{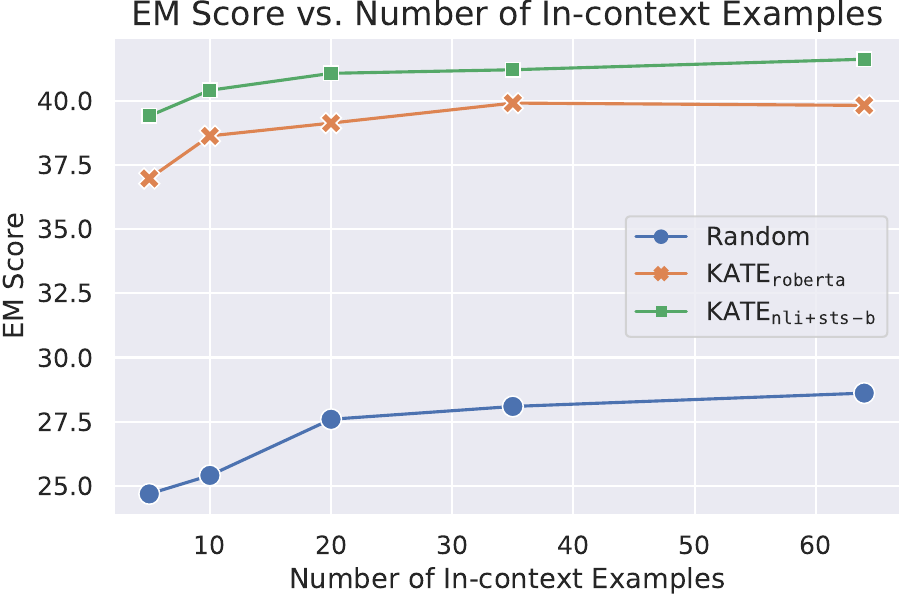}
    \includegraphics[width=\columnwidth]{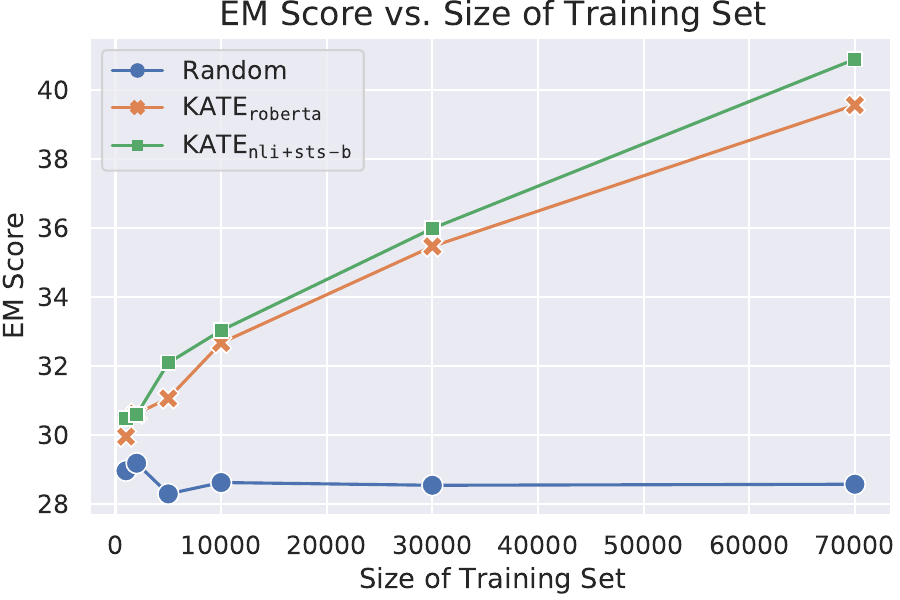}
    \caption{Left: Ablation study on the effect of number of in-context examples for GPT-$3$ for different selection methods. Right: Ablation study on the effect of the size of training set for retrieval on KATE. Two representative sentence encoders are used in the ablation study.}
    \label{fig:ablation_2_figs}
\end{figure*}

We also evaluate KATE on the open-domain QA tasks, as shown in Table~\ref{tab:QA_result}.
For the QA tasks, we compare with some state-of-the-art methods such as RAG~\cite{lewis2020retrieval} and T5~\cite{raffel2019exploring}. Both methods require fine-tuning on the specific datasets.
The KATE method again improves GPT-$3$'s few-shot prediction accuracies substantially across various benchmarks.
It is worth noting that the fine-tuned transformer models serve as better sentence encoders for retrieval purpose (compared with the RoBERTa-large model without fine-tuning).
KATE$_{\text{nli}}$ and KATE$_{\text{nli+sts-b}}$ improve upon KATE$_{\text{roberta}}$ because this time fine-tuning on NLI or STS-B datasets is helpful for retrieving semantically similar questions from the QA datasets.
Moreover, on the NQ and TriviaQA datasets, further fine-tuning on the STS-B dataset improves KATE's results.
We also try reducing the number of in-context examples to be as small as five for both the random and KATE methods, where KATE outperforms the baseline as well.
Therefore, the advantage of KATE over the random baseline holds for both small and large numbers of in-context examples.
More details can be found in Section~\ref{sec:number}.
We evaluate the other baseline $k$NN$_{\text{roberta}}$ by using the top-$1$ nearest neighbor.
We also explore using 64 nearest neighbors (10 for TriviaQA) to determine the answer (by majority voting explained in Section~\ref{sec:baseline_method}). The EM score tends to be similar to retrieving the top-$1$ nearest neighbor.
These $k$NN baseline results again suggest that the retrieval module and GPT-$3$ work together to achieve better performance.

To investigate why the retrieval examples are helpful, we further present a case study. Concretely, the retrieved in-context examples from the NQ dataset are shown in Table~\ref{tab:NQ_retrieval_eg}.
For the first and second cases, the random baseline provides wrong answers because GPT-$3$ is unable to recall the exact detail.
However, the in-context examples selected by KATE contain the correct details, which facilitates GPT-$3$ to answer the questions.
For the third test question, the random baseline leads GPT-$3$ to misinterpret the question as asking for a specific location.
In contrast, KATE selects similar questions which ask for the origins of objects. Using these in-context examples, GPT-$3$ is able to interpret and answer the question correctly.
\vspace{-1mm}
\section{Analysis and Ablation Study}
\vspace{-2mm}
\subsection{Number of In-context Examples}
\label{sec:number}
We first investigate the impact of the number of in-context examples on KATE's performance.
Concretely, on the NQ dataset, we choose the number of in-context examples to be $5$, $10$, $20$, $35$, and $64$, and KATE$_{\text{nli+sts-b}}$ is compared with the random baseline and KATE$_{\text{roberta}}$ across different settings.
As shown in the left plot of Figure~\ref{fig:ablation_2_figs}, both KATE and the random baseline benefit from utilizing more in-context examples.
However, KATE consistently outperforms the random selection method, even when the number of in-context examples is as few as $5$.
This result is interesting because in practice, employing less in-context leads to more efficient inference with GPT-$3$.
\vspace{-2mm}
\subsection{Size of Training Set for Retrieval}
\vspace{-1mm}
We further examine how the size of the training set may influence the KATE method.
On the NQ dataset, we create new subsets from the original training set, with sizes of 1k, 2k, 5k, 10k, 30k, and 70k, respectively.
In-context examples are retrieved from these subsets instead of the original training set.
The number of nearest neighbors is set to 64.
We compare KATE$_{\text{nli+sts-b}}$ with the random selection method and KATE$_{\text{roberta}}$, and the results are shown in the right plot of Figure~\ref{fig:ablation_2_figs}.
For KATE$_{\text{roberta}}$ and KATE$_{\text{nli+sts-b}}$, as the size of the training set for retrieval increases, the EM scores also increase.
In contrast, the result of the random sampling baseline does not change much.
Intuitively, as the training size gets larger, it is more likely for KATE to retrieve relevant in-context examples to help GPT-$3$ answer a question correctly.
As we have shown previously in Table~\ref{tab:NQ_retrieval_eg}, the retrieved in-context examples could provide critical detailed information to GPT-$3$, thus helping GPT-$3$ to better answer the questions.
\begin{table}[ht]
\vspace{+2mm}
\centering
\begin{small}
\begin{tabular}{@{}c||cccccc@{}}
\toprule[1.2pt]
\textbf{Trial}    & 1    & 2    & 3    & Default & Reverse\\ \hline
\textbf{EM Score} & 42.0 & 42.5 & 42.0 & 41.6 & 42.8   \\ \bottomrule[1.2pt]
\end{tabular}%
\caption{Ablation study on the effect of in-context example orders for GPT-$3$ on the NQ dataset using KATE$_\text{nli+sts-b}$. For the default order, the example $A$ is to the left of example $B$ if $A$ is closer to the test prompt $x$ than $B$ in the embedding space. For the reverse order, the example A is to the right of example B.}
\vspace{-2mm}
\label{tab:ablation_example_order}
\end{small}
\end{table}
\subsection{Order of In-context Examples}
\vspace{0.5mm}
Moreover, we explore how the order of in-context examples may affect KATE's results.
As mentioned in Section~\ref{sec:KATE_proposal_method}, under the standard setting, the retrieved in-context examples are ordered such that $d(x_i, x) \leq d(x_j, x)$ whenever $i < j$.
Here, we randomly permute the order of in-context examples in the NQ dataset for the proposed KATE$_{\text{nli+sts-b}}$ method, and conduct the experiments for $3$ different orders.
Additionally, we explore the reverse order where $d(x_i, x) \leq d(x_j, x)$ whenever $i < j$.
The results are presented in Table~\ref{tab:ablation_example_order}.
On this particular NQ dataset, the reverse order performs the best.
One possible explanation is that since tokens next to each other have similar positional embeddings, putting the most similar sentences close to the test example may be helpful for GPT-3 to leverage the corresponding information.
However, we also did the experiments on the WQ and TriviaQA and find that the default order performs slightly better than the reverse order.
Hence, the choice of orders is data-dependent.
Addtionally, it can be observed that the variation among the NQ results tends to be quite small (compared with the difference between the random baseline and KATE), indicating that the example order does not have a significant impact on KATE's performance.

\section{Related Work}
\paragraph{Pre-trained Language Models}
NLP systems have made tremendous progress by pre-training models on unlabeled text.
For text classification tasks, notable models include BERT~\cite{devlin2018bert}, RoBERTa~\cite{liu2019roberta}, and XLNet~\cite{yang2019xlnet}.
For text generation tasks, notable models include BART~\cite{lewis2019bart}, T5~\cite{raffel2019exploring}, mT5~\cite{xue2020mt5}, XLM~\cite{lample2019cross}, GPT~\cite{radford2018improving}, and GPT-2~\cite{radford2019language}.
These models encapsulate rich information to facilitate a wide range of downstream tasks ranging from natural language understanding to generation.
These models can be adapted to many different tasks via fine-tuning.
GPT-$3$~\cite{brown2020language}, however, can be adapted to many downstream tasks without fine-tuning.
Given just a few in-context examples, GPT-$3$ is able to quickly pick up patterns and produce answers analogously both in terms of the answer style and content.
Thus, GPT-$3$ may be considered as a pattern recognizer to perform in-context learning.
People have just started trying to understand GPT-$3$ from different perspectives.
As mentioned in the introduction, ~\cite{hendrycks2020measuring} studies which categories of questions GPT-$3$ is more capable of answering.
Our work focuses on how to choose good in-context examples.

\paragraph{Retrieval-based Text Generation}
There is a long history of applying information retrieval in text generation~\cite{sumita1991experiments}.
It is very related to the exemplar-based learning~\cite{jakel2008generalization, ziyadi2020example}.
The central idea is to treat retrieved samples as exemplars/prototypes and perform some editings on them.
Some representative applications in the field of deep learning include machine translation~\cite{gu2018search}, sentiment transfer~\cite{li2018delete, guu2018generating}, QA~\cite{karpukhin2020dense, mao2020generation}, dialogue generation~\cite{yan2016learning, cai2018skeleton, song2016two, pandey2018exemplar, weston2018retrieve, wu2019response}, text summarization~\cite{cao2017faithful, peng2019text}, data-to-text generation~\cite{peng2019text}, and text-to-code generation~\cite{hashimoto2018retrieve}.
However, all these retrieve-and-edit frameworks require their decoders to be trained from scratch.
This makes the editor network task- and data-specific.
In contrast, GPT-$3$ in one perspective can be regarded naturally as a universal editor, adaptive to a wide range of tasks.
Our work uniquely examines how to maximize the advantage of using GPT-$3$ without fine-tuning.
For example, the more semantically similar context we provide to GPT-$3$, the better results the model can generate.
Other editors or generators do not have this ability.

\paragraph{Improve NLP Systems with $k$NN}
A recent line of work tries to incorporate nonparametric methods to improve a given model's performance.
These methods first access the test sample's hidden representation and look for the nearest neighbors of this test sample in the database.
Once the nearest neighbors are found, their labels are used to augment the model's prediction.
For example, the newly introduced $k$NN-LM~\cite{khandelwal2019generalization}, $k$NN-MT~\cite{khandelwal2020nearest}, and BERT-$k$NN~\cite{kassner2020bert} generate the next token by retrieving the nearest $k$ neighbors from the datastore.
Another related work is $k$NN classification model~\cite{rajani2020explaining}, where they use $k$NN as backoff when the confidence is low from the fine-tuned classification model.
There are two key differences between our work and other approaches.
First, other approaches modifies the model's next token distribution using the nearest $k$ neighbors.
However, we only changes the conditional text using the nearest $k$ neighbors.
Second, other approaches can access the model's parameters and embeddings which we do not have access to.
Instead, we use some other independently pre-trained models to get the sentence embeddings to retrieve nearest $k$ neighbors.

\section{Conclusion}
This work presented a first step towards investigating the sensitivity of GPT-$3$ to in-context examples.
To this end, we proposed KATE, a non-parametric selection approach 
that retrieves in-context examples according to their semantic similarity to the test samples.
On several natural language understanding and generation tasks, the proposed method improves GPT-$3$'s performance, over the random sampling baseline, by a significant margin.
Moreover, we found that fine-tuning the sentence embeddings for retrieval on task-related datasets gave rise to further empirical gains.
Detailed ablation studies were conducted to explore the robustness of KATE to different hyperprameters, such as the number of in-context examples, examples' order, \emph{etc}.
We hope this work could provide insights for better understanding the behaviors of GPT-$3$ and represents a helpful step towards further improving its few-shot capabilities.






\bibliography{naaclhlt2019}
\bibliographystyle{acl_natbib}

\end{document}

%% file: macro.tex
\usepackage{booktabs}
\usepackage{graphicx}
\usepackage{multirow}

\usepackage{textcomp,xspace}    

\usepackage{url}
\usepackage{algorithm}
\usepackage{amsfonts}
\usepackage{algpseudocode}
\usepackage{soul}
\usepackage{makecell}
\input{math_commands.tex}

\algnewcommand{\LineComment}[1]{\State \(\triangleright\) #1}

\newcommand{\upv}{\vspace{-.0cm}}
\newcommand{\downv}{\vspace{-.0cm}}

\definecolor{gred}{RGB}{219,68,55}
\definecolor{gblue}{RGB}{66,133,244}
\definecolor{gyellow}{RGB}{244,180,0}
\definecolor{ggreen}{RGB}{15,157,88}
\definecolor{ggrey}{RGB}{115,115,115}

\newcommand{\colorR}[1]{\textcolor{gred}{\textbf{#1}}}
\newcommand{\colorG}[1]{\textcolor{ggreen}{\textbf{#1}}}
\newcommand{\colorB}[1]{\textcolor{gblue}{#1}}

%% file: math_commands.tex
\usepackage{amsmath,amsthm,amsfonts,amssymb}
\usepackage{bm}
\usepackage{graphicx}
\usepackage{sidecap}
\usepackage{mathrsfs}
\usepackage{multirow, multicol}
\usepackage{caption}

\usepackage{wrapfig}
\usepackage{booktabs}

\usepackage{natbib} 
\def\1{\bm{1}}

\def\vs{{\bm{s}}}

\def\vv{{\bm{v}}}

\def\vx{{\bm{x}}}
\def\vy{{\bm{y}}}

\DeclareMathAlphabet{\mathsfit}{\encodingdefault}{\sfdefault}{m}{sl}
\SetMathAlphabet{\mathsfit}{bold}{\encodingdefault}{\sfdefault}{bx}{n}



%% file: ToTTo_retrieval_eg.tex
\begin{table*}[ht]
        \centering
        \setlength{\tabcolsep}{11pt}
        \def\arraystretch{1.05}
        \resizebox{0.75\textwidth}{!}{
        \begin{tabular}{@{}c||l@{}}
\toprule[1.5pt]
\multirow{2}{*}{\textbf{Test Table}} & \textbf{Table}: \textless{}page\_title\textgreater Trey Johnson  \textless{}section\_title\textgreater College  \textless{}table\textgreater \textless{}cell\textgreater 32 \textless{}col\_header\textgreater  \\
& GP \textless{}cell\textgreater 4.8 \textless{}col\_header\textgreater RPG  \textless{}cell\textgreater 2.3 \textless{}col\_header\textgreater APG   \textless{}cell\textgreater 23.5 \textless{}col\_header\textgreater PPG \\
\hline
& \textbf{Table}: \textless{}page\_title\textgreater Dedric Lawson  \textless{}section\_title\textgreater College  \textless{}table\textgreater \textless{}cell\textgreater 9.9 \textless{}col\_header\textgreater \\
& RPG \textless{}cell\textgreater 3.3  \textless{}col\_header\textgreater APG  \textless{}cell\textgreater 19.2 \textless{}col\_header\textgreater PPG \\ \multirow{1}{*}{\textbf{Retrieved}} & \textbf{Sentence}: Dedric Lawson averaged 19.2 points, 9.9 rebounds and 3.3 assists per game. \\
\multirow{1}{*}{\textbf{Examples}} & \textbf{Table}: \textless{}page\_title\textgreater Carsen Edwards  \textless{}section\_title\textgreater College  \textless{}table\textgreater \textless{}cell\textgreater 3.8 \textless{}col\_header\textgreater \\
& RPG  \textless{}cell\textgreater 2.8 \textless{}col\_header\textgreater APG  \textless{}cell\textgreater 18.5 \textless{}col\_header\textgreater PPG \\ & \textbf{Sentence}: Edwards averaged 18.5 points, 3.8 rebounds and 2.8 assists per game. \\
\hline
\multirow{3}{*}{\textbf{Predictions}} & \textbf{Ground-truth}: Trey Johnson averaged 23.5 \underline{points}, 4.8 \underline{rebounds}, and 2.3 \underline{assists} in 32 games. \\
& \textbf{Random}: Trey Johnson averaged 23.5 points per game in his senior year at the University of Texas. \\
& \textbf{KATE}: Johnson averaged 23.5 \underline{points}, 4.8 \underline{rebounds} and 2.3 \underline{assists} per game. \\ \bottomrule[1.5pt]
\end{tabular}%
        }
    \upv
    \caption{A sample of retrieved in-context examples from the ToTTo dataset. The first block shows a table string from the test set. The second block provides two retrieved examples. The last block includes the ground-truth sentence and sentences by the random selection and KATE methods. For the KATE method, GPT-$3$ pays more attention to detailed information such as the number of points, rebounds, and assists. In contrast, the random selection method leads GPT-$3$ to generate details which do not exist in the original table.}
    \label{tab:ToTTo_retrieval_eg}
    \downv
    \end{table*}

%% file: NQ_retrieval_eg.tex
\begin{table*}[ht]
        \centering
        \setlength{\tabcolsep}{10pt}
        \def\arraystretch{1.05}
        \resizebox{\textwidth}{!}{
        \begin{tabular}{@{}l|l@{}}
            \toprule[1.5pt]
            \multicolumn{1}{c|}{\textbf{In-Context Examples}} & \multicolumn{1}{c}{\textbf{Predictions}} \\
            \midrule
            \multicolumn{2}{c}{\textbf{Question}: The Mughal Gardens of Rashtrapati Bhavan is modelled on which garden?}\\
            \midrule
            The Mughal Garden of Rashtrapati Bhavan is modelled on?  \underline{The \colorG{Persian} style of architecture} & \textbf{Ground-truth}: \colorB{Persian garden} \\
            Who built the first Mughal Garden in India?  \underline{Babur} & \textbf{KATE}: \colorG{The Persian gardens}\\
            The landscape design of the Gardens of Versailles is known as which style? \underline{French garden} & \textbf{Random Baseline}: \colorR{Shalimar gardens}\\
             
            \midrule
            \multicolumn{2}{c}{\textbf{Question}: What city was Zeus the patron god of?}\\
            \midrule
            What is the symbol of Zeus the Greek God?  \underline{Bull} &  \textbf{Ground-truth}: \colorB{Olympia} \\
            Where did Zeus spend most of his time? \underline{Mount Olympus} & \textbf{KATE}: \colorG{Olympia} \\
            Where was the statue of Zeus at \colorG{Olympia} located? \underline{In the Temple of Zeus} &  \textbf{Random Baseline} \colorR{Athens}\\
            
            \midrule
            \multicolumn{2}{c}{\textbf{Question}: Where did the Dewey decimal system come from?} \\
            \midrule
            Where did the formula for area of a circle come from? \underline{Archimedes} & \textbf{Ground-truth}: \colorB{Melvil Dewey} \\
            Where did the name jack russell come from? \underline{Reverend John Russell} & \textbf{KATE}: \colorG{Melvil Dewey} \\ 
            Where did the letters of the alphabet come from?  \underline{The Phoenician alphabet} & \textbf{Random Baseline}: \colorR{the library of Congress} \\
            
            \bottomrule[1.5pt]
        \end{tabular}
        }
    \upv
    \caption{Samples of retrieved in-context examples from the NQ dataset. Each block begins with a test question. Three retrieved Q-A pairs are shown on the left. Predictions by the KATE method and useful details from in-context examples are shown in \colorG{Green}. Gold-standard references are shown in \colorB{Blue}. Predictions by the random selection method are shown in \colorR{Red}.}
    \label{tab:NQ_retrieval_eg}
    \downv
    \end{table*}